\documentclass[runningheads]{llncs}

\usepackage{eccv}

\usepackage{eccvabbrv}

\usepackage{graphicx}
\usepackage{booktabs}
\usepackage{fontawesome}

\usepackage[accsupp]{axessibility}  

\usepackage{hyperref}

\usepackage{orcidlink}

\usepackage{multirow}

\usepackage{pifont}
\begin{document}

\title{Open-Vocabulary SAM: Segment and Recognize Twenty-thousand Classes Interactively}

\titlerunning{Open-Vocabulary SAM}

\author{Haobo Yuan\inst{1}
\orcidlink{0000-0001-9770-7720}
\and
Xiangtai Li\inst{1}
\orcidlink{0000-0002-0550-8247}
\and
Chong Zhou\inst{1}
\orcidlink{0000-0002-9776-7739}
\and
Yining Li\inst{2}
\orcidlink{0000-0003-1753-0887}
\and
Kai Chen\inst{2}
\orcidlink{0000-0002-6820-2325}
\and
Chen Change Loy\inst{1}
\orcidlink{0000-0001-5345-1591}
}

\authorrunning{H. Yuan et al.}

\institute{
{\small S-Lab, Nanyang Technological University \and Shanghai AI Laboratory}\\
Project Page: \url{https://www.mmlab-ntu.com/project/ovsam}\\
E-mail: \email{yuanhaobo@whu.edu.cn}, \email{xiangtai94@gmail.com}
}
\maketitle

\begin{abstract}
The CLIP and Segment Anything Model (SAM) are remarkable vision foundation models (VFMs). SAM excels in segmentation tasks across diverse domains, whereas CLIP is renowned for its zero-shot recognition capabilities. This paper presents an in-depth exploration of integrating these two models into a unified framework. Specifically, we introduce the Open-Vocabulary SAM, a SAM-inspired model designed for simultaneous interactive segmentation and recognition, leveraging two unique knowledge transfer modules: SAM2CLIP and CLIP2SAM. The former adapts SAM's knowledge into the CLIP via distillation and learnable transformer adapters, while the latter transfers CLIP knowledge into SAM, enhancing its recognition capabilities. Extensive experiments on various datasets and detectors show the effectiveness of Open-Vocabulary SAM in both segmentation and recognition tasks, significantly outperforming the na\"{i}ve baselines of simply combining SAM and CLIP. Furthermore, aided with image classification data training, our method can segment and recognize approximately 22,000 classes. 

\end{abstract}

\newcommand{\cavan}[1]{{\color{blue}(cavan: {#1})}} 
\newcommand{\lxt}[1]{{\color{cyan}(xiangtai: {#1})}} 
\newcommand{\chong}[1]{{\color{magenta}(chong: {#1})}} 

\section{Introduction}
\label{sec:intro}

\begin{figure*}
    \centering
    \includegraphics[width=1.\textwidth]{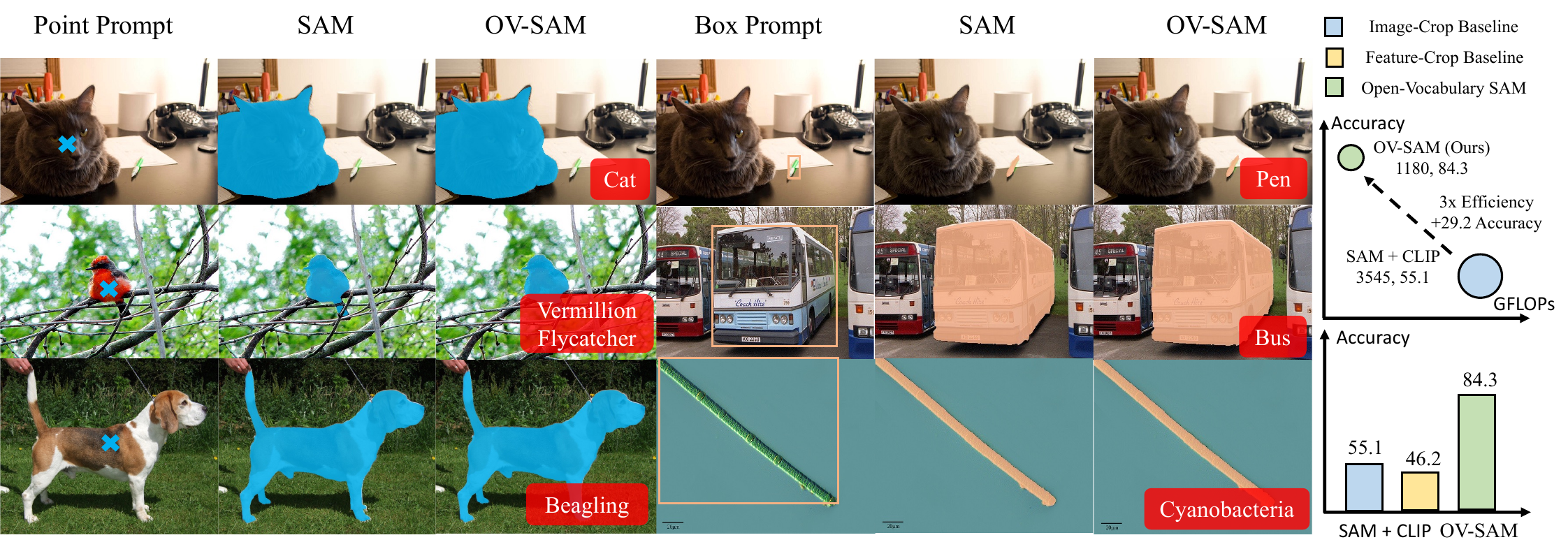}
    \caption{Open-Vocabulary SAM not only can segment anything with prompts just like SAM but also has the capability of recognition in the real world, like CLIP. With drastically lower computational cost, Open-Vocabulary SAM has a higher recognition performance than directly combining SAM and CLIP with image or feature cropping (measured on the COCO open vocabulary benchmark).}
    \label{fig:teaser_01}
    \vspace{-4mm}
\end{figure*}

The Segment Anything Model (SAM)~\cite{kirillov2023segment} and CLIP~\cite{radford2021learning} have made significant strides in various vision tasks, showcasing remarkable generalization capabilities in \textit{segmentation} and \textit{recognition}, respectively. SAM, in particular, has been trained with a massive dataset of mask labels, making it highly adaptable to a wide range of downstream tasks through interactive prompts. On the other hand, CLIP's training with billions of text-image pairs has given it an unprecedented ability in zero-shot visual recognition. This has led to numerous studies~\cite{gu2021open, zang2022open, wu2023towards, xu2022simple} exploring the extension of CLIP to open vocabulary tasks, such as detection and segmentation.

\begin{figure*}[t]
    \centering
    \includegraphics[width=1.\textwidth]{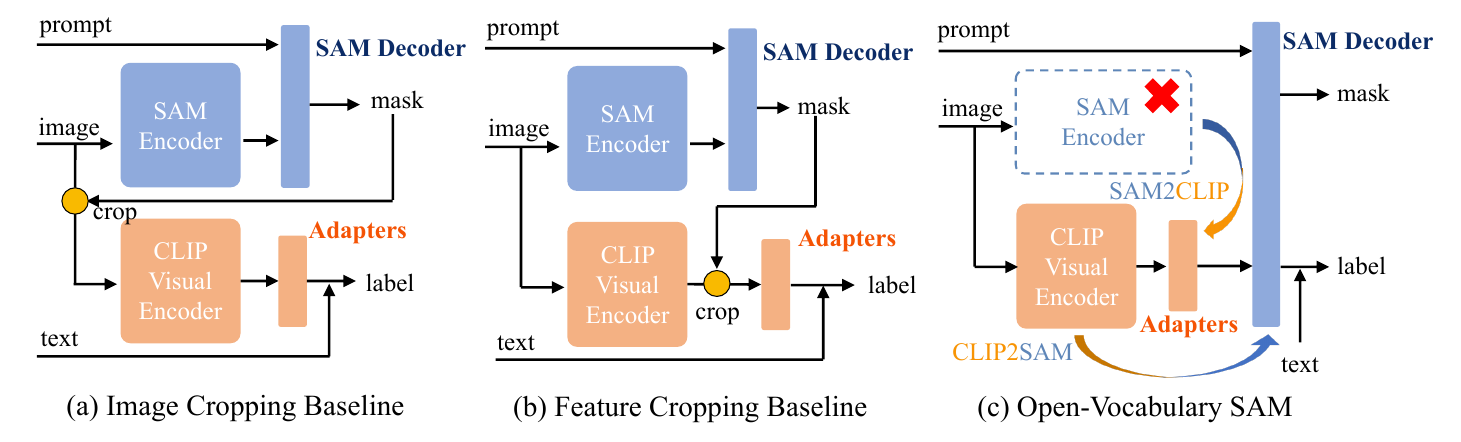}
    \caption{Comparison of two simple SAM-CLIP combination baselines (a) and (b), and our proposed single encoder architecture (c). The adapters for (a) and (b) are optional and can be replaced with various designs (please refer to Sec.~\ref{sec:main_results} for details). Note that, in our method, the SAM encoder will be discarded during inference.}
    \label{fig:teaser_02}
    \vspace{-3mm}
\end{figure*}

While SAM and CLIP offer considerable advantages, they also have inherent limitations in their original designs. SAM, for instance, lacks the capability to recognize the segments it identifies. Efforts to overcome this by integrating a classification head have been made~\cite{li2023semantic,zhao2023fast}, but these solutions are constrained to specific datasets or closed-set settings. On the other hand, CLIP, which is trained using image-level contrastive losses, faces challenges in adapting its representations for dense prediction tasks. To address this, several studies~\cite{wu2023baron, yu2023fcclip, wu2023clipself, shi2023edadet, gu2021open} have investigated ways to align CLIP's representation for dense predictions. However, these approaches tend to be dataset-specific and not universally applicable. For example, some research has focused on open vocabulary segmentation on the ADE-20k~\cite{zhou2019semantic} dataset, using the COCO~\cite{coco_dataset} dataset for pre-training. Merging SAM and CLIP in a na\"{i}ve manner, as illustrated in Fig.~\ref{fig:teaser_02} (a) and (b), proves to be inefficient. This approach incurs substantial computational expenses and yields subpar results, including recognition of small-scale objects, as evidenced by our experimental results.

In this study, we address these challenges with a unified encoder-decoder framework that integrates a CLIP encoder and a SAM decoder, as depicted in Fig.~\ref{fig:teaser_02} (c). To bridge these two distinct components effectively, we introduce two novel modules, \textit{SAM2CLIP} and \textit{CLIP2SAM}, facilitating dual knowledge transfer. First, we distill knowledge from the SAM encoder to the CLIP encoder using \textit{SAM2CLIP}. This distillation process is uniquely executed not directly on the CLIP encoder, which is kept frozen to maintain its existing knowledge, but rather on a lightweight transformer-like adapter using a pixel-wise distillation loss. The adapter takes multi-scale features as input, with the goal of aligning CLIP features with SAM representation. On the decoding side, the \textit{CLIP2SAM} module transfers knowledge from the frozen CLIP encoder to the SAM decoder. In particular, we design a feature pyramid adapter with a RoIAlign operator to be jointly trained with the SAM decoder. Both modules are lightweight and naturally combine the strengths of CLIP and SAM.

Following the spirit of SAM, we enhance our model's recognition capabilities by harnessing the power of established semantic datasets, including COCO~\cite{coco_dataset}, LVIS~\cite{gupta2019lvis}, and ImageNet-22k~\cite{deng2009imagenet}. This strategy elevates our model to the versatility of SAM, endowing it with enhanced capability to segment and recognize any objects, as shown in Fig.~\ref{fig:teaser_01}. As our approach is an adaptation of SAM, it is flexible enough to be integrated with various detectors, making it suitable for both closed-set and open-set environments.

We conduct extensive experiments across a range of datasets and scenarios, encompassing closed-set and open-vocabulary interactive segmentation. Notably, when compared to basic combined baselines, our approach demonstrates superior performance, achieving over 2\% improvement in IoU and 3\% in mAP with various detectors on the COCO dataset. In particular, in the case of recognition on LVIS, our approach achieves over 20\% improvements over previous adapters.   Furthermore, by expanding our approach with a more diverse array of datasets, we have developed a versatile, interactive tool suitable for practical applications. For detailed results, we direct the reader to Sec.~\ref{sec:exp} and the appendix.

\section{Related Work}
\label{sec:related_work}



\noindent
\textbf{Vision Language Models (VLMs).} Vision-language pre-training has given rise to models with aligned image and text representations~\cite{JayaramanG14, jia2021scaling, VILT, radford2021learning,li2022scaling}. Recent studies on contrastive vision-language pre-training~\cite{jia2021scaling, radford2021learning, LIT, EVA-CLIP} have significantly improved the generalization ability of recognition models. Meanwhile, several works~\cite{li2022blip,ALBEF,VILT,li2023blip} aim to design better optimization goals for downstream multi-modal tasks, including caption and visual question answering. Among these works, CLIP models~\cite{radford2021learning} that are pre-trained on billion-scale image-text pairs have shown impressive zero-shot classification performance on a wide range of datasets. Our goal is to enable SAM to perform recognition tasks with the help of pre-trained VLMs.

\noindent
\textbf{Open Vocabulary Dense Prediction.}  This direction aims to recognize region visual concepts of arbitrary categories described by texts, which includes object detection~\cite{zareian2021open,gu2021open, wu2023towards, wu2023betrayed,xie2023mosaicfusion}, semantic segmentation~\cite{ xu2023side,zhou2022maskclip,li2023transformer,zhou2023rethinking, Li2022SFNetFA, sfnet}, and panoptic segmentation~\cite{xu2023open,yu2023fcclip,yang2023panoptic}. This necessitates the alignment between region and text representations with the help of VLMs~\cite{jia2021scaling,radford2021learning,EVA-CLIP}. For open-vocabulary detection, a series of works~\cite{gu2021open, zang2022open, wu2023baron, shi2023edadet} distill knowledge from the CLIP models to recognize novel objects. In contrast to distillation-based methods, several works~\cite{FVLM, wu2023cora} directly build object detectors upon frozen CLIP CNNs. For open-vocabulary segmentation, the typical works~\cite{xu2022simple, ding2022decoupling, xu2023open, yu2023fcclip} first generate class-agnostic mask proposals and then classify the proposals with CLIP. Recently, several works~\cite{xu2023open, yu2023fcclip} build the mask generator upon the frozen diffusion model~\cite{rombach2022high} and CLIP model~\cite{radford2021learning}. Meanwhile, several studies~\cite{qi2022open,qi2022fine,Maaz2022Multimodal,kim2021oln} focus on class-agnostic segmentation and detection to enrich generalization ability in various domains. However, most approaches are trained and tested on specific datasets. Our approach is based on SAM, which provides a general, interactive tool to support different open vocabulary detectors.

\noindent
\textbf{Prompting in Computer Vision.} 
Prompting, originating from in-context learning in natural language processing (NLP) as seen in works like Brown \etal~\cite{brown2020language} and Rubin \etal~\cite{rubin2021learning}, leverages a large language model to infer unseen tasks through context-specific input-output pairs. Recent studies~\cite{bar2022visual, Painter, SegGPT, zhou2022coop, fang2023explore, balavzevic2023towards, liu2023effective} have explored in-context learning for visual tasks. Common techniques involve mask image modeling~\cite{MaskedAutoencoders2021, beit, yu2022point-bert} for cross-task visual prompting, as employed by approaches like Painter~\cite{Painter} and Bar \etal~\cite{bar2022visual}. SAM~\cite{kirillov2023segment} demonstrates in-context learning through interactive segmentation, using diverse visual prompts like points, boxes, and masks, although it is limited to class-agnostic mask prediction. Meanwhile, other studies~\cite{gao2021clip, Lian_2022_SSF, chen2022vitadapter, hu2021lora, omgseg} have concentrated on efficient parameter tuning of visual foundation models, typically focusing on a single model. Our work uniquely bridges two models, CLIP and SAM, exploring their combined potential for enhanced general segmentation and recognition capabilities. In particular, we adopt visual prompts (box, point) as the model's inputs.

\noindent
\textbf{Segmentation Anything Model.} SAM~\cite{kirillov2023segment} presents a new data engine and portable model for segmentation. Subsequent research has employed SAM as an interactive segmentation tool for various vision tasks, including grounding~\cite{liu2023grounding}, tracking~\cite{cheng2023segmenttracking}, distillation~\cite{mobile_sam, zhou2023edgesam}, medical analysis~\cite{wu2023medical,chen2023ma}, and generation~\cite{zhang2023personalize}. While some studies use SAM and CLIP for segmentation~\cite{chen2023semantic,han2023boosting,wang2023sam,li2024clipsam,pan2023tokenize}, none have yet integrated VLMs and SAM into a unified model capable of both segmentation and recognition of novel classes. Our work makes the first attempt to merge the capabilities of VLMs with SAM for enhanced task versatility.

\section{Methodology}
\label{sec:method}

\begin{figure*}[t!]
    \centering
   \includegraphics[width=1.\textwidth]{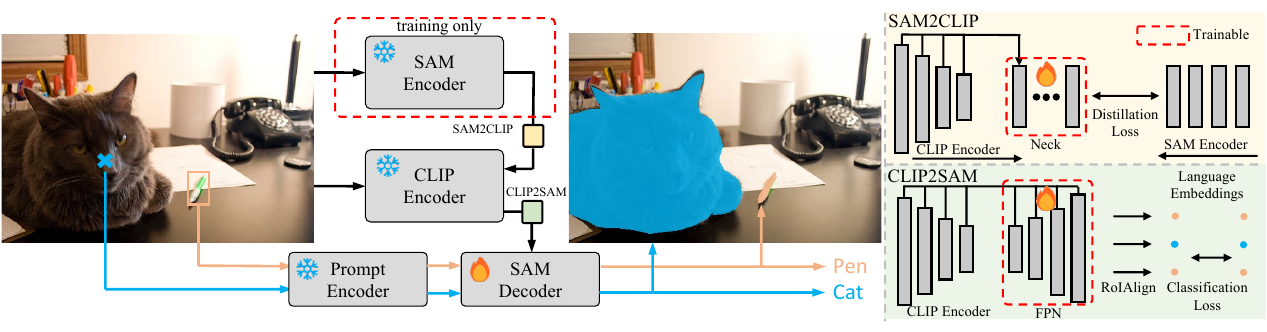}
    \caption{Illustration of Open-Vocabulary SAM. For training, the SAM encoder is as a teacher network, while SAM2CLIP plays the role of a student network and aligns the knowledge of SAM into CLIP. The CLIP2SAM transfers the CLIP knowledge to the SAM decoder and performs joint segmentation and classification for close-set and open vocabulary settings. }
    \label{fig:method}
\end{figure*}

We first review the SAM, CLIP, and combined baselines in Sec.~\ref{sec:method_pre}. Then, we detail our Open Vocabulary SAM in Sec.~\ref{sec:method_open_sam}. Last, we present our model's training details and application in Sec.~\ref{sec:method_train_application}.

\subsection{Preliminaries and Baselines}
\label{sec:method_pre}

\noindent
\textbf{SAM.} SAM is a prompt-driven segmentor. It contains an image encoder, a prompt encoder, and a light-weight mask decoder. Here, we use box prompts as an example. We denote an input image as $X \in \mathbb{R}^{H\times {W}\times 3}$ and input visual prompts as $P \in \mathbb{R}^{N \times 4}$, where ${H}\times {W}$ are the spatial size, $N$ is the number of box prompts. The image encoder is a modified vision transformer (ViT). It encodes an image into dense feature $F_{SAM} \in \mathbb{R}^{ \frac{H}{16}  \times \frac{W}{16} \times d}$. The prompt encoder encodes $P$ into sparse prompts $Q_{sp}$. Meanwhile, mask tokens $Q_{mask}$ and an IoU token $Q_{IoU}$ are initialized for the mask decoder.

The mask decoder takes the image feature $F$, sparse prompts $Q_{sp}$, mask tokens $Q_{mask}$, and the IoU token $Q_{IoU}$ as input. All the inputs will be concatenated and encoded with a lightweight two-way transformer. Consequently, each mask token is transformed into a dynamic linear classifier, capable of calculating the foreground mask probability for every sparse prompt. Simultaneously, the IoU token is tasked with predicting the confidence score for each mask. 
Considering the multi-granular nature of SAM's data annotations, encompassing both instance and part level, $Q_{mask}$ naturally encodes multi-granularity. 
Our study concentrates exclusively on the object level, which aligns more closely with prevalent real-world applications and datasets such as COCO~\cite{coco_dataset} and LVIS~\cite{gupta2019lvis}.

\noindent
\textbf{CLIP.}
Given an input image $X$ and a corresponding caption $C$, the CLIP framework processes these modalities to produce respective embeddings: the image embedding $E_{I}$, derived from its image encoder, and the text embedding $\mathbf{t}$, obtained from its text encoder. In the context of open-vocabulary object detection and segmentation, CLIP's capability to generalize beyond fixed class labels is leveraged to replace traditional classifiers. For instance, in open-vocabulary detection scenarios, the text embedding $\mathbf{t_c}$ for the $c$-th object category is generated by inputting the category name into the CLIP text encoder. This process can employ a single template prompt, such as "a photo of \{category\}," or multiple prompt templates. Subsequently, for a given region embedding $r$, that is produced by the RoI-Align~\cite{he2017mask}, the classification score for the $c$-th category is computed as: $p_c = \frac{\text{exp}(\tau \cdot <\mathbf{r}, \mathbf{t_c}>)}{\sum_{i=0}^{C} \text{exp}(\tau \cdot <\mathbf{r}, \mathbf{t_i}>)}$,
where $<\cdot,\cdot>$ denotes the cosine similarity, and $\tau$ is a learnable or fixed temperature to re-scale the value.

\noindent
\textbf{Combined Baselines.} 
We introduce two different baselines for combining CLIP and SAM, as depicted in Fig.~\ref{fig:teaser_02} (a) and (b). The first approach, termed the `cropped image baseline', employs the SAM mask decoder's output to segment and resize the original input image. This processed image then serves as the input for the CLIP image encoder, and, in conjunction with the CLIP text embedding, the mask is classified using the similarities between visual and text embeddings. The second approach referred to as the `cropped CLIP image feature baseline', employs the same initial CLIP feature extraction step. However, in this method, masks predicted by the SAM decoder are used to crop the CLIP image features. Subsequent pooling of these masked features yields the final label, akin to baseline (a).

While both baselines enable zero-shot inference of images, they exhibit a noticeable knowledge gap on specific datasets. To address this, we draw inspiration from recent advancements in visual prompting or adapters~\cite{zhou2022coop,chen2022vitadapter}. Specifically, we propose incorporating additional learnable tokens as an adapter to fine-tune the model for enhanced performance on downstream datasets. These zero-shot inference capabilities and the fine-tuned models constitute our primary comparison baselines under various experimental conditions, detailed in Sec.~\ref{sec:main_results}.

\subsection{Open Vocabulary SAM} 
\label{sec:method_open_sam}
While both baseline models can be enhanced through visual prompting or adapters, as we will discuss in Sec.~\ref{sec:exp}, they face several challenges in real-world applications. First, the requirement for two independent backbones in the combined model increases computational costs ($Prob.1$). Second, SAM and CLIP are trained with distinct objectives – SAM through supervised learning and CLIP via contrastive learning – and there is limited research on knowledge transfer between such diverse architectures ($Prob.2$). Third, despite adapter integration, significant performance gaps remain in recognizing small objects ($Prob.3$). Fourth, there is a lack of exploration into integrating open-vocabulary capabilities for SAM and CLIP, particularly in the context of feature fusion and data scaling ($Prob.4$). Our work aims to solve these problems in a unified yet effective framework.

\noindent
\textbf{Unified Architecture.} We design a unified architecture for both segmentation and recognition to address $Prob.1$.
Specifically, we adopt the frozen CLIP visual encoder as our feature extractor. Then, both SAM's mask decoder and prompt encoder are appended behind the CLIP encoder.
The meta-architecture of open-vocabulary SAM is shown in Fig.~\ref{fig:teaser_02} (c), with the more detailed version shown in Fig.~\ref{fig:method}. This unified architecture is made possible via the \textbf{SAM2CLIP}, which transfers knowledge of SAM to CLIP with distillation, and \textbf{CLIP2SAM}, which employs CLIP knowledge and combines the SAM mask decoder for recognition. We have chosen convolution-based visual backbones for the frozen CLIP backbone, aligning with previous studies that have highlighted their superiority in capturing spatial structures~\cite{FVLM,xu2023dst-det}. The efficacy of different CLIP backbones is further explored in Sec.~\ref{sec:exp_ablation_analysis}.

\noindent
\textbf{SAM2CLIP.}
To resolve $Prob.2$, we design the SAM2CLIP module that bridges the gap in feature representations learned by SAM and CLIP, using adaptation and distillation methods. Through comprehensive experiments, we discovered that employing distillation loss $L_{distill}$ along with transformer-based adapters~\cite{ViT}, yields effective results. Specifically, the distillation process involves a simple pixel-wise approach, where SAM-Huge serves as the teacher, and the frozen CLIP equipped with an adapter assumes the student's role. We then implement a per-pixel mean squared error (MSE) loss to align the SAM feature $F_{sam}$ with the CLIP feature $E_{I}$, as detailed below:
\begin{align}
    L_{distill} = \mathrm{MSE}(F_{sam}, E_{I}).
\end{align}
We design a multi-scale adapter $A_{sam2clip}$ to align the features from CLIP and SAM. In particular, we take pyramid CLIP features $E_{I}^{i}, i=1, 2, 3$ as the inputs. Such pyramid features contain both high-resolution and semantic information, which is proven crucial for semantic segmentation~\cite{kirillov2019panoptic}. The MSE loss is revised as follows:
\begin{align}
    L_{distill} = \mathrm{MSE}(F_{sam}, A_{sam2clip}(\mathrm{Fusion}(E_{I}^{i}))),
    \label{eq:sam2clip}
\end{align}
where $A_{sam2clip}$ comprises several transformer layers, and $\mathrm{Fusion}$ is achieved by bilinear upsampling and addition.

With SAM2CLIP, we can even achieve comparable segmentation results with the SAM-Huge with much lower computational costs. As detailed in Sec.~\ref{sec:exp_ablation_analysis}, we observe that employing convolution-based methods specifically designed for backbone adaptation~\cite{MIMDet,chen2022vitadapter} results in sub-optimal outcomes. The reason for this might be in the inherent architecture of the SAM encoder, which is purely based on ViT. A symmetrical structure is crucial for effective knowledge transfer. With the implementation of SAM2CLIP, we can achieve segmentation results comparable to those of SAM-Huge across various detectors, while significantly reducing computational costs.

\noindent
\textbf{CLIP2SAM.} 
This module aims to leverage CLIP's knowledge to enhance the recognition capabilities of the SAM decoder. A straightforward approach involves appending a label token $Q_{label}$ to the existing mask token $Q_{mask}$ and IoU token $Q_{IoU}$. Using $Q_{label}$, we introduce a specialized adapter to facilitate the transfer of knowledge from the frozen CLIP to the SAM decoder. Subsequently, the enhanced $Q_{label}$, combined with the output of the prompt encoder and adapted CLIP features, is fed into a two-way transformer. Following the cross-attention process, the improved $Q_{label}$ undergoes further refinement through a multilayer perceptron (MLP), ensuring better alignment with CLIP's text embedding. The final labels are derived by calculating the distance between the refined label token and the CLIP text embedding.

This design, however, falls short of recognizing small objects ($Prob.3$) since the adaptation only involves the single-scale feature, which is mainly focused on segmentation. We present a simple yet effective solution to handle this issue, introducing a lightweight feature pyramid network (FPN) for CLIP2SAM adaption. As shown in Fig.~\ref{fig:method}, the pyramid network extracts multi-scale CLIP features as the inputs. Then, we apply the RoI-Align~\cite{he2017mask} operation to extract region features. Like the R-CNN framework~\cite{he2017mask},  we apply one convolution layer and a MLP to learn the feature embedding without introducing cross-attention in the mask decoder. 
In particular, for point prompts, we first obtain the corresponding masks via the SAM decoder and obtain the box via the corresponding masks.  For box prompts, we can directly send it to the FPN for region feature extraction. Given that our method incorporates only a few convolution layers, it does not significantly increase computational costs compared to the original SAM.

\noindent
\textbf{Open Vocabulary.} 
To tackle $Prob.4$, the open-vocabulary challenge, we leverage the knowledge embedded in the frozen CLIP backbone, which aids in recognizing novel and unseen objects during inference. In line with previous studies~\cite{FVLM,wu2023clipself}, we fuse the learned class scores with those from the frozen CLIP via a geometric mean to leverage information from both the CLIP and CLIP2SAM. Additionally, we investigate various strategies to expand the vocabulary size, such as joint training with multiple datasets, as detailed in Sec.~\ref{sec:exp_ablation_analysis}. Our experimental results show that the model scales effectively with large datasets.

While our approach can address the open vocabulary challenge, it is important to distinguish the setting of Open-Vocabulary SAM from that of previous open vocabulary segmentation methods~\cite{chen2023semantic,huynh2022open, ding2023open, xu2023masqclip}. Unlike previous techniques, which typically depend on a separate segmentor to produce mask proposals, our method only uses visual prompts, such as boxes and points, to generate masks and labels. Furthermore, our method specifically targets foreground objects, aligning with the ImageNet~\cite{deng2009imagenet} datasets and the purpose of box prompts. To show the effectiveness of our method, we compare the performance of our method with previous open vocabulary segmentation methods in Sec.~\ref{sec:main_results}.

\subsection{Training and Application}
\label{sec:method_train_application}

\noindent
\textbf{Training and Loss Function.} 
We first use the SA-1B (1\%) dataset~\cite{kirillov2023segment} for training the SAM2CLIP module to transfer SAM's knowledge into open-vocabulary SAM, with the loss $L_{distill}$ (Equ.~(\ref{eq:sam2clip})). Then, we joint train the CLIP2SAM and mask decoder using segmentation mask and label annotations from COCO or LVIS. The final loss function is given as $L = \lambda_{cls}L_{t\_cls} + \lambda_{ce}L_{t\_ce} + \lambda_{dice}L_{t\_dice}$. Here, $L_{t\_ce}$ is the Cross-Entropy (CE) loss for mask classification, and $L_{t\_ce}$ and $L_{t\_dice}$ are mask Cross Entropy (CE) loss and Dice loss~\cite{dice_loss} for segmentation, respectively. In addition, we adopt joint training with the ImageNet dataset for classifying over 22,000 categories.

\noindent
\textbf{Inference and Demo Tools.} Our model performs inference like SAM, with points and boxes as visual prompts. Specifically, we test boxes and points as visual prompts for the encoder in Sec.~\ref{sec:exp}. In the appendix, we show a demo of our model, which can segment and recognize with prompts. 
\section{Experiments}
\label{sec:exp}

\noindent
\textbf{Datasets and Metrics.} We mainly use COCO~\cite{coco_dataset} and LVIS~\cite{gupta2019lvis} datasets for the experiments. Moreover, we also use part of SA-1B~\cite{kirillov2023segment} data (1\%) for SAM2CLIP knowledge transfer. For COCO, we report the results of both close-set and open-vocabulary settings for the instance segmentation task. In particular, following Zareian~\etal~\cite{zareian2021open}, we split 48 base classes with annotations and 17 target classes without annotations. We use the base class annotations for training. For LVIS datasets, we adopt the open-vocabulary setting and report the results of $AP_{rare}$ for novel classes. For evaluation metrics, we report the accuracy of recognition for reference to evaluate the recognition capability. Meanwhile, each prompt's intersection-over-union (IoU) with its ground truth mask is also adopted to evaluate the segmentation ability of our method. 
As mentioned in Sec.~\ref{sec:method_open_sam}, different from previous open vocabulary segmentation tasks, boxes or points serve as visual prompts in our method.
 
\noindent
\textbf{Baselines.} As shown in Fig.~\ref{fig:teaser_02} (a) and (b), based on different adapter designs, we append these adapters to the different locations of the combined models. For example, when using CoOp~\cite{zhou2022coop}, we append the learnable tokens by combining them with CLIP features. For several convolution-based adapters~\cite{chen2022vitadapter}, we add the extra convolution layers along with SAM or CLIP backbone for fair comparison. By default, we adopt SAM-huge and CLIP R50x16. 

\begin{table*}[t!]
    \centering
    \caption{Comparison of combined baselines and Open-Vocabulary SAM using visual prompts. ``*'' indicates using mask center point as prompts, while others indicate using ground truth boxes prompts. $IoU_b$ and $IoU_n$ refer to the average IoU for each mask of base classes and novel classes, respectively.}
    \resizebox{\textwidth}{!}{
    \begin{tabular}{c|ccc|ccc|ccc}
    \toprule[0.2em]
    \multirow{2}{*}{Method} & \multicolumn{3}{c|}{COCO}  & \multicolumn{3}{c|}{LVIS} &  \multirow{2}{*}{FLOPs} & \multirow{2}{*}{\#Param}
    \\
    & $IoU_{b}$ & $IoU_{n}$ & Acc &$IoU_{b}$ & $IoU_{n}$ & Acc\\
    \midrule
    Image-Crop baseline & 78.1 & 81.4 & 46.2 & 78.3 & 81.6 & 9.6 & 3,748G & 808M \\
    Feature-Crop baseline & 78.1 & 81.4 & 55.1 & 78.3 & 81.6 & 26.5 & 3,545G & 808M  \\
    \hline
    Image-Crop baseline + CoOp~\cite{zhou2022coop} & 79.6 & 82.1 & 62.0 & 80.1 & 82.0 & 32.1 & 3,748G & 808M  \\
    Feature-Crop baseline + CoOp~\cite{zhou2022coop} & 79.6 & 82.1 & 70.9 & 80.1 & 82.0 & 48.2 & 3,545G & 808M \\
    \hline
    Open-Vocabulary SAM & 81.5 & 84.0 & 84.3 & 80.4 & 83.1 & 66.6 & 1,180G & 304M  \\
    \hline
    Image-Crop baseline* & 60.7 & 66.7 & 24.5 & 53.0 & 62.3 & 6.2 & 3,748G & 808M \\
    Feature-Crop baseline* & 60.7 & 66.7 & 32.1 & 53.0 & 62.3 & 11.0 & 3,545G & 808M  \\
    \hline
    Image-Crop baseline + CoOp~\cite{zhou2022coop}* & 64.7 & 66.7 & 28.2 & 58.9 & 64.2 & 8.3 & 3,748G & 808M \\
    Feature-Crop baseline + CoOp~\cite{zhou2022coop}* & 64.7 & 66.7 & 35.1 & 58.9 & 64.2& 13.2 & 3,545G & 808M  \\
    \hline
    Open-Vocabulary SAM* & 68.4  &  65.2 & 76.7 & 63.6  & 67.9 & 60.4  & 1,180G & 304M  \\
    \bottomrule[0.2em]
    \end{tabular}
    }
    \label{tab-results:gt_box_prompt_testing}
\end{table*}

\begin{table*}[t!]
    \centering
    \caption{Comparison of combined baselines and Open-Vocabulary SAM on prompts generated by the open vocabulary detector. For the LVIS dataset, only `normal' and `frequent' classes are in the training set. The labels are generated by each baseline or our method. We adopt Detic~\cite{zhou2022detecting} as the OV-Detector to provide box prompts.}
    \resizebox{\linewidth}{!}{
    \begin{tabular}{c|ccc|cccc|cc}
    \toprule[0.2em]
    \multirow{2}{*}{Method} & \multicolumn{3}{c|}{COCO}  & \multicolumn{4}{c|}{LVIS} &  \multirow{2}{*}{FLOPs} & \multirow{2}{*}{\#Params}
    \\
    & $AP_{base}$ & $AP_{novel}$ & $AP$ & $AP_{rare}$ & $AP_{norm}$ &  $AP_{freq}$ & $AP$ \\
    \midrule
    Image-Crop baseline + CoOp~\cite{zhou2022coop} & 26.2 & 31.2 & 27.3 & 19.8 &	18.3 &	16.3 & 17.2 & 3,748G & 808M  \\
    Feature-Crop baseline + CoOp~\cite{zhou2022coop} & 28.0 & 33.8 & 29.5 & 24.2 &	21.4 &	18.6 & 20.8 & 3,545G & 808M \\
    Open-Vocabulary SAM & 31.1 & 36.0 & 32.4 & 24.0 & 21.3&22.9&22.4 & 1,180G & 304M \\
    \bottomrule[0.2em]
    \end{tabular}
    }
    \label{tab-results:ov-detector}
\end{table*}

\noindent
\textbf{Implementation Details.} We implement our models in PyTorch~\cite{paszke2019pytorch} with MMDetection~\cite{chen2019mmdetection}. We use 8 A100 GPUs for distributed training. Each mini-batch has two images per GPU. The optimizer is AdamW~\cite{loshchilov2018decoupled} with a weight decay of 0.0001. We adopt full image size for a random crop in the pre-training and training process following Cheng~\etal~\cite{cheng2021mask2former}. All the class names are transferred into CLIP text embedding, following previous works~\cite{gu2021open}. We train each model for 12 epochs for fair comparison. Due to the limitation of computation costs, we do not adopt joint SAM data and COCO data training. We first perform training the SAM2CLIP on SA-1B (1\%), and then we finetune the model on COCO or LVIS data. Please refer to the supplementary material for more details.

\noindent
\textbf{Benchmark Setup.} Open Vocabulary SAM is different from both SAM~\cite{kirillov2023segment} and open vocabulary segmentation methods~\cite{huynh2022open, ding2023open, xu2023masqclip}. Compared with Open-Vocabulary SAM, SAM~\cite{kirillov2023segment} cannot recognize segmented objects while open vocabulary segmentation methods~\cite{huynh2022open, ding2023open, xu2023masqclip} usually rely on a separate segmentor (e.g., Mask-RCNN~\cite{he2017mask}) instead of prompts like points or boxes. For a complete evaluation of Open-Vocabulary SAM, we set the benchmark as three parts, including 1).comparison with combined baselines to verify the effectiveness; 2).comparison with open-vocabulary segmentation methods to evaluate the recognition; 3).comparision with SAM models to evaluate segmentation.

\begin{table*}[t]
    \centering
    \caption{Comparison of Open-Vocabulary SAM with previous methods on open vocabulary instance segmentation. The results presented in the table all use the Mask-RCNN~\cite{he2017mask} with ResNet-50 backbone. Different from previous works, Open-Vocabulary SAM uses the bounding box as the prompt to generate the mask. We report mask AP50 following base-novel setting as~\cite{huynh2022open}.}
    \resizebox{0.65\linewidth}{!}{
    \begin{tabular}{c|c|cc|ccc}
    \toprule[0.2em]
    \multirow{2}{*}{Method}& \multirow{2}{*}{Venue} & \multicolumn{2}{c|}{Constrained}  & \multicolumn{3}{c}{Generalized}
    \\
    && Base & Novel & Base & Novel & All\\
    \midrule
    XPM~\cite{huynh2022open}& CVPR'22 & 42.4 & 24.0 & 41.5 & 21.6 & 36.3\\
    MaskCLIP~\cite{ding2023open} & ICML'23 & 42.8 & 23.2 & 42.6 & 21.7 & 37.2\\
    MasQCLIP~\cite{xu2023masqclip} & ICCV'23 & 40.9 & 30.1 & 40.7 & 28.4 & 37.5\\
    \hline
    Open-Vocabulary SAM & (Ours) & 41.7 & 37.5 & 39.3 & 39.8 & 39.4\\
    \bottomrule[0.2em]
    \end{tabular}
    }
    \label{tab:ov_segs}
\end{table*}

\begin{table*}[t!]
    \centering
    \caption{Comparison of mask quality with various detectors on COCO dataset. We report the mask mean AP for comparison. The masks are generated by each method, while the labels are from the corresponding detectors.}
    \resizebox{0.95\linewidth}{!}{
    \begin{tabular}{cc|ccc|ccc|ccc}
    \toprule[0.2em]
    Method & Detectors & mAP & AP50 & AP75 & APS & APM & APL  & \#Params & FLOPs \\
    \midrule
    SAM-Huge & Faster-RCNN (R50) & 35.6 & 	54.9 &	38.4 & 17.2 & 39.1 & 51.4 & 641M & 3,001G  \\
    SAM-Huge (finetuned) & Faster-RCNN (R50) & 35.8 & 55.0 & 38.4 & 16.5 & 38.6 & 53.0 & 641M & 3,001G \\
    Open-Vocabulary SAM  & Faster-RCNN (R50) & 35.8	& 55.6 &	38.3 & 16.0 & 38.9 & 53.1 & 304M & 1,180G \\
    \hline
    SAM-Huge &  Detic (swin-base) & 36.4 & 57.1 & 39.4 & 21.4 & 40.8 & 54.6 & 641M & 3,001G  \\
    SAM-Huge (finetuned) & Detic (swin-base) & 36.8 & 57.4 & 39.8 & 20.8 & 40.6 & 55.1 & 641M & 3,001G  \\
    Open-Vocabulary SAM  & Detic (swin-base) & 36.7 & 57.2 & 39.7 & 20.7 & 40.8 & 54.9 & 304M & 1,180G \\
    \hline
    SAM-Huge &  ViTDet (Huge) & 46.3 & 72.0 & 49.8 & 25.2 & 45.5 & 59.6 & 641M & 3,001G  \\
    SAM-Huge (finetuned) & ViTDet (Huge) & 46.5 & 72.3 & 50.3 & 25.2 & 45.8 & 60.1 & 641M & 3,001G  \\
    Open-Vocabulary SAM  & ViTDet (Huge) & 48.8 & 73.8 & 52.9 & 24.8 & 46.3 & 64.2 & 304M & 1,180G \\
    \bottomrule[0.2em]
    \end{tabular}
    }
    \label{tab-res-SAM}
\end{table*}

\begin{table}[t!]
\centering
\caption{Scaling up with large-scale datasets.}
\resizebox{0.75\linewidth}{!}{
\begin{tabular}{l|c|cc}
\toprule[0.2em]
Datasets & Accuracy & \#vocaulary & \#images \\
\midrule
LVIS & 83.1 & 1,203 &  99K\\
V3Det & 78.7 & 13,204 & 183K\\
I-21k & 44.5 & 19,167 & 13M \\
V3Det + LVIS & 82.7 & 13,844 & 282K \\
V3Det + LVIS + I-21k & 83.3 & 25,898 & 13M \\
V3Det + LVIS + I-21k + Object365 & 83.0 & 25,970 & 15M \\
\bottomrule[0.2em]
\end{tabular}
}
\label{tab-results:scaleup}
\end{table}

\begin{table}[t!]
    \centering
    \caption{The effectiveness of each component. We use Detic~\cite{zhou2022detecting} as the detector. The labels are generated by the corresponding model. S2C and C2S denote SAM2CLIP and CLIP2SAM respectively. The baseline refers to the image-crop variant.}
    \resizebox{0.75\linewidth}{!}{
    \begin{tabular}{c|cccccc}
    \toprule[0.2em]
    Setting & $AP$ & $AP_{base}$ & $AP_{novel}$ & FLOPs (G) & \#Params (M) \\
    \midrule
    Baseline + CoOp~\cite{zhou2022coop} & 29.5 & 28.0 & 33.8 & 3,545 & 808  \\
    \hline
    Our + S2C & 28.7 & 27.3 & 33.3 & 1,127 & 291 \\
    Our + S2C + C2S & 34.4 & 33.1 & 38.0 & 1,180 & 304 \\
    \bottomrule[0.2em]
    \end{tabular}
    }
    \label{tab-ablation:effectiveness}
\end{table}

\subsection{Main Results}
\label{sec:main_results}
\noindent
\textbf{Comparison with Combined Baselines Using Ground Truth.} To avoid the influence of other modules, we first demonstrate the recognition ability of our model in Tab.~\ref{tab-results:gt_box_prompt_testing}. Compared to the simple combined approaches, adding adapters with joint co-training leads to better results. However, the recognition ability is still limited on both COCO and LVIS. Our Open-Vocabulary SAM achieves the best results on both boxes and points as visual prompts. We observe more significant gains on LVIS datasets. We argue that LVIS contains more small objects, which is more challenging than COCO. Our method can solve $Prob.2$ and lead to over 20\% accuracy improvement. Although the segmentation quality is pretty good (about 80 IoU on COCO and LVIS with box prompt), our method still achieves 2\% IoU improvements. This indicates the effectiveness of our joint co-training on mask prediction and classification. Compared with boxes as prompts, using points as prompts is more challenging since the location clues of points are much weaker than boxes. However, our approach is still better than combined baselines or them with adapters.

\noindent
\textbf{Comparison with Combined Baselines on OV-Detector.} In Tab.~\ref{tab-results:ov-detector}, we adopt a more challenging setting by using the box prediction from the existing open-vocabulary detector to simulate the interactive segmentation process with deviation. We choose the representative Detic~\cite{zhou2022detecting} as the open-vocabulary detector. Again, our method also achieves the best performance on both COCO and LVIS datasets. In particular, on COCO, compared with previous works~\cite{zhou2022coop}, our method achieves 3.0 mask mAP improvements with much lower costs. 

\noindent
\textbf{Comparison with Open-vocabulary Segmentation.} In Tab.~\ref{tab:ov_segs}, we compare our method with previous open-vocabulary instance segmentation methods. Open-Vocabulary SAM uses the boxes generated by Mask-
RCNN~\cite{he2017mask} and previous works use the masks of Mask-RCNN. The results show that our Open-Vocabulary SAM performs strongly, especially in the novel classes.

\noindent
\textbf{Comparison with SAM on Various Detectors.} In Tab.~\ref{tab-res-SAM}, we test the mask prediction quality of our model and original SAM on different detectors. Our method performs better than the original SAM and performs comparably with fine-tuned SAM. It is worth noting that our Open-Vocabulary SAM has much lower computational costs and parameters than SAM.

\noindent
\textbf{Comparison on Interactive Segmentation.} In Tab.~\ref{tab-results:sams} (right), we compare interactive segmentation performance. Notably, our approach excels beyond interactive segmentation and can recognize classes in an open-vocabulary setting.

\noindent
\textbf{Visualization Comparison.} In Fig.~\ref{fig:visual_comparison_fig}, we compare our approach with the feature-crop baseline. Our model shows better performance in classifying small and rare object classification and handling occlusion scenarios.

\noindent
\textbf{Model as a Zero Shot Annotation Tool.} In addition to COCO and LVIS standard datasets training, following the spirit of SAM, we also scale up our model by training it with more data (Tab.~\ref{tab-results:scaleup}). In particular, we adopt more detection data (V3Det~\cite{wang2023v3det}, Object365~\cite{shao2019objects365}) and classification data (ImageNet22k~\cite{deng2009imagenet}). Owing to significant costs, we have not conducted comparisons with other baselines for this setting. Rather, we have adapted our method into an interactive annotation tool capable of segmenting and recognizing over 22,000 classes.

\subsection{Ablation Studies and Analysis}
\label{sec:exp_ablation_analysis}

\begin{table*}[!t]
    \footnotesize
	\centering
    \caption{Ablation studies on COCO open-vocabulary dataset. We use boxes of the ground truth as a prompt to generate masks and labels. (left): Ablation on SAM2CLIP design. (right): Ablation on CLIP2SAM design.}
    \label{tab-ablation:SAM2CLIP_design}
    \label{tab-ablation:CLIP2SAM_design}
    \resizebox{0.45\linewidth}{!}{
        \begin{tabular}{c|ccccc}
        \toprule[0.2em]
        Setting & IoU & FLOPs (G) & \#Params (M) \\
        \midrule
        SAM & 78.7 & 3,001 & 641  \\
        Conv-L + MultiScale Neck & 78.3 & 1,313  & 321  \\
        Conv-L + SingleScale Neck & 73.6 & 1,280  & 307  \\
        R50x16 + MultiScale Neck & 78.1 & 1,180 & 304  \\
        R50 + MultiScale Neck & 77.3  & 728 & 165 \\
        \bottomrule[0.2em]
        \end{tabular}
    }\hfill
    \resizebox{0.54\linewidth}{!}{
        \begin{tabular}{l|ccccccc}
        \toprule[0.2em]
        Setting & IoU & Acc & $AP_{base}$ & $AP_{novel}$ \\
        \midrule
        SAM2CLIP (baseline) & 78.1 & 54.2 & 27.6 & 33.2 &  \\
        \hline
        + Cls Token & 81.3 & 79.3 & 29.8 & 34.5  \\
        + Cls Token \& CLIP MLP fusion & 80.9 & 78.9 & 29.0 & 33.9   \\
        + light FPN (Ours) & 81.5 & 84.3 & 31.1 & 36.0  \\
        \bottomrule[0.2em]
        \end{tabular}
    }
\end{table*}

\begin{table*}[!t]
    \footnotesize
	\centering
    \caption{(left): Ablation study on different CLIP backbone. We test results on the COCO open-vocabulary dataset. We use boxes of the ground truth as a prompt to generate masks and labels. (right): Comparison with other SAM models.}
    \label{tab-results:sams}
    \label{tab-ablation:differnt_clip}
    \resizebox{0.47\linewidth}{!}{
        \begin{tabular}{c|cccccc}
        \toprule[0.2em]
        Backbone & IoU & Acc & \#FLOPs(G) & \#Params (M) \\
        \midrule
        RN50 & 77.3 & 50.8 & 728 & 165 \\ 
        RN50x16 & 78.1 & 55.1 & 1,180 & 304  \\
        RN50x64 & 78.1 &  54.1 & 2,098 & 568    \\
        ConvNeXt-L & 78.3 & 59.1 & 1,313  &  321  \\
        ViT-L-14 & 38.6 & 14.3 & 2,294 & 441\\
        \bottomrule[0.2em]
        \end{tabular}
    }\hfill
    \resizebox{0.47\linewidth}{!}{
        \begin{tabular}{l|c|cc}
        \toprule[0.2em]
        Method & 1-IoU (COCO) & cls. & open.\\
        \midrule
        SAM~\cite{kirillov2023segment}~(H) & 78.2 & - & - \\
        SEEM~\cite{zou2023segment}~(T) & 73.7 & \checkmark & -\\
        Semantic-SAM~\cite{li2023semantic}~(T) & 76.1  & \checkmark & - \\
        OV-SAM (ours) & 81.7  & \checkmark & \checkmark \\
        \bottomrule[0.2em]
        \end{tabular}
    }
\end{table*}

\noindent
\textbf{Effectiveness of SAM2CLIP and CLIP2SAM.} We first verify the effectiveness of our proposed two modules in Tab.~\ref{tab-ablation:effectiveness}. We adopt image-crop variant of the baseline for comparison.
In particular, by sharing a single backbone, we observe a significant drop in the number of parameters and FLOPs, with a little drop in segmentation performance. The slight drop is caused by the domain gap between SAM data and COCO data during the training of the SAM2CLIP module. However, after adding our CLIP2SAM module and joint co-training with mask classification and prediction, a significant improvement in both segmentation and classification is observed, with just a negligible increase in compute cost.

\noindent
\textbf{Detailed Design on SAM2CLIP.} In Tab.~\ref{tab-ablation:SAM2CLIP_design}, we explore the detailed design of SAM2CLIP in the first stage of open-vocabulary SAM training. The results show that distillation benefits most when multi-scale features are adopted, suggesting that both high-resolution features and high-level semantics are important to align CLIP's feature with the SAM's feature. 

\begin{figure}[t]
    \centering
   \includegraphics[width=0.75\textwidth]{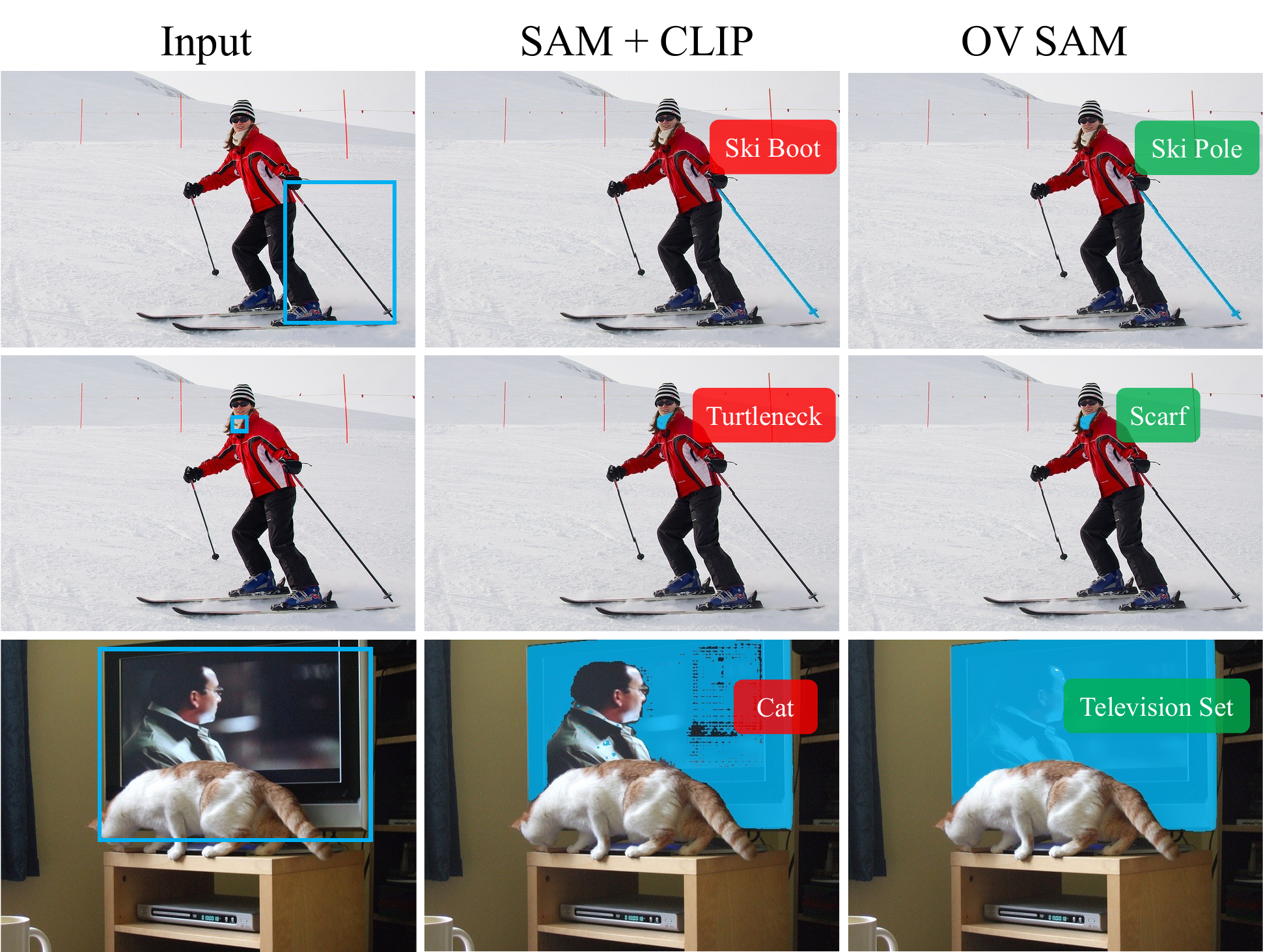}
    \caption{Visualization Comparison. We compare the mask and classification results of the image-crop baseline (SAM + CLIP) and Open-Vocabulary SAM (OV SAM). The predicted labels are presented on the mask.}
    \label{fig:visual_comparison_fig}
\end{figure}

\begin{figure}[t]
    \centering
   \includegraphics[width=\textwidth]{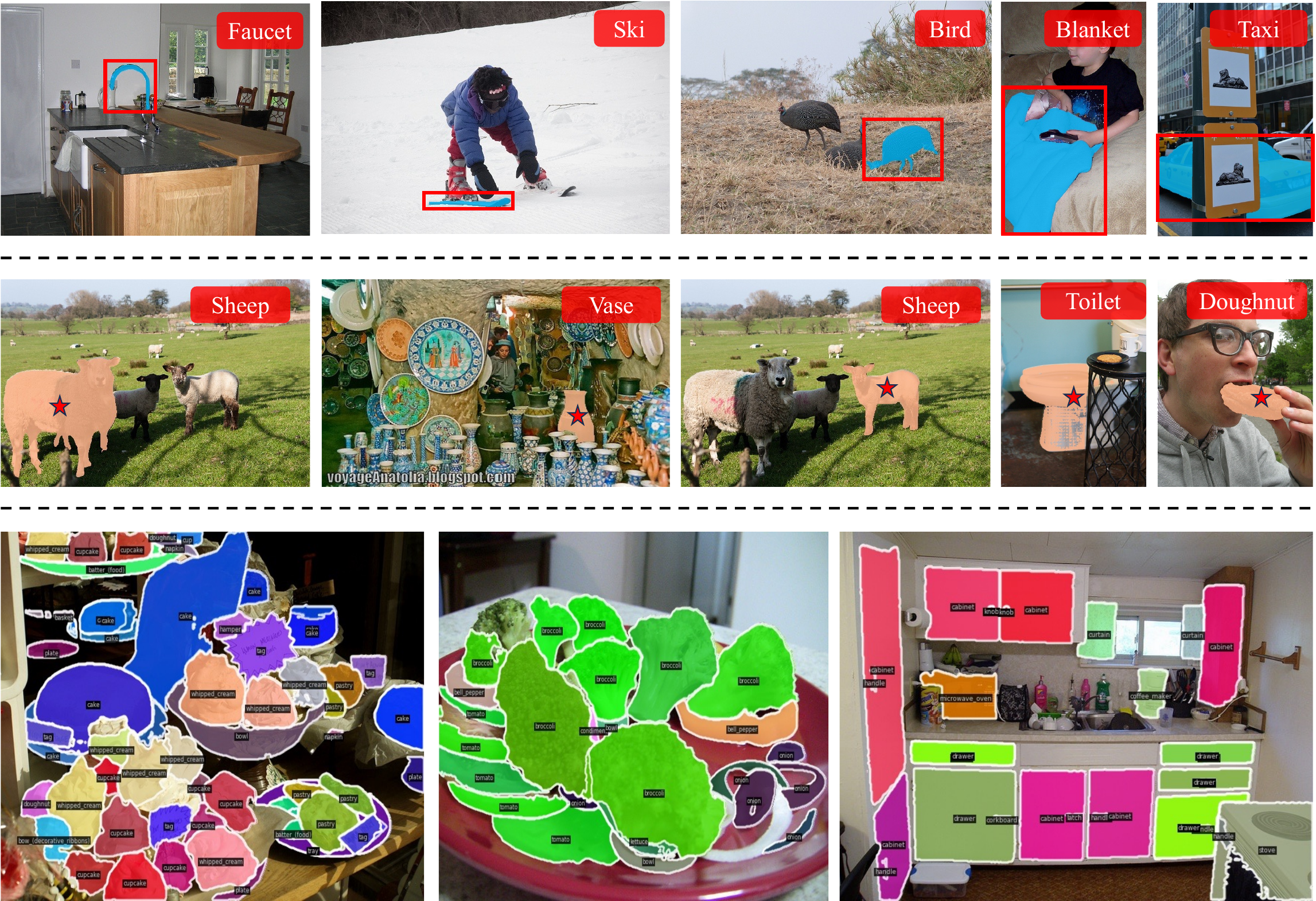}
    \caption{Qualitative results of Open-Vocabulary SAM. In the visualization, boxes refer to box prompts (up), and the red stars refer to the point prompts (middle). The masks can be merged into a segmentation map (bottom). We show the mask and labels generated by the proposed Open-Vocabulary SAM. Our method can segment and recognize open vocabulary objects in diverse scenes identified by prompts.}
    \vspace{-2em}
    \label{fig:visualization_results}
\end{figure}

\noindent
\textbf{Detailed Design on CLIP2SAM.} In Tab.~\ref{tab-ablation:CLIP2SAM_design}, we present extensive design for the CLIP2SAM module. We compare two designs: a simple classification token with cross attention (Cls Token) and a combination of this token with mask pooled CLIP feature (CLS Token \& CLIP MLP fusion). These designs work better than the combined baseline shown in the first row. Nonetheless, due to resolution constraints, these variants cannot handle small objects well, as shown in Fig.~\ref{fig:visual_comparison_fig}. In contrast, our design improves the performance considerably.

\noindent
\textbf{Ablation on Different CLIP Backbones.} In Tab.~\ref{tab-ablation:differnt_clip}, we explore the effect of frozen CLIP visual backbone. We do not add the CLIP2SAM. Motivated by recent works~\cite{FVLM,yu2023fcclip,xu2023dst-det,wu2023clipself}, CNN-based CLIPs encapsulate more structural information, which is good for our goal since we have location-sensitive visual prompts. Thus, adopt CNN-based CLIPs, aligned with previous works. As shown in the table, we find ConvNext large achieves the best performance, but we choose RN50x16 since it has comparable performance and better efficiency.

\section{Conclusion}
\label{sec:conclusion}
We present Open Vocabulary SAM, a SAM-inspired method for interactive segmentation and recognition. Unlike previous open-vocabulary detection and segmentation methods, our method explores interactive open-vocabulary segmentation for the first time. Given the user's inputs, such as boxes or points, the proposed approach can interactively segment and label each visual prompt. Compared with the combined baselines and various visual adapters, our proposed CLIP2SAM and SAM2CLIP are both efficient and effective in various settings. Our open vocabulary segmentation is compatible with different detectors, including open-vocabulary detectors and close-set detectors. With more data, our model plays a similar role as SAM, offering an effective annotation tool for both segmentation and instance labeling. In particular, our method can perform large vocabulary segmentation and recognition over 22K classes. We hope our Open-Vocabulary SAM can provide a solid baseline for combining the strengths of different forms of vision foundation models and inspire further research.

\appendix
\begin{center}{
    \bf \Large Appendix
}
\end{center}

\renewcommand{\thetable}{S\arabic{table}}
\renewcommand{\thefigure}{S\arabic{figure}}
\renewcommand{\thesection}{S\arabic{section}}
\setcounter{figure}{0}
\setcounter{table}{0}
\noindent
\textbf{Overview.} In the supplementary, we provide more details and results to support our main paper. The contents are presented as follows: we first present more details and discussions of our method in Sec.~\ref{sec:more_relatd_work}. Then, we report more results in Sec.~\ref{sec:more_exp_results}. Finally, we present a demo and discuss future work in Sec.~\ref{sec:demo}. Please refer to our GitHub repo at \url{https://github.com/HarborYuan/ovsam} for more details about the demo and implementation.

\begin{table}[b]
   \centering
    \caption{Comparison with Recent Joint SAM and CLIP models~\cite{chen2023semantic,wang2023sam,zhang2023recognize, han2023boosting}.}
\resizebox{\linewidth}{!}{
   \begin{tabular}{r | c c c c c}
      \toprule
     Property & {\footnotesize SSA\cite{chen2023semantic}} & SAM-CLIP\cite{wang2023sam} & RecognizeAnything\cite{zhang2023recognize} & Sambor\cite{han2023boosting} & Ours \\        \hline
        Single Backbone & \raisebox{-0.5ex}{\ding{55}} &  \raisebox{-0.5ex}{\checkmark} &  \raisebox{-0.5ex}{\ding{55}} & \raisebox{-0.5ex}{\ding{55}} &  \raisebox{-0.5ex}{\checkmark}  \\
        Object-Level Classification& \raisebox{-0.5ex}{\ding{55}} & \raisebox{-0.5ex}{\ding{55}} & \raisebox{-0.5ex}{\ding{55}} & \raisebox{-0.5ex}{\checkmark} & \raisebox{-0.5ex}{\checkmark}\\
        Interactive Segmentation &  \raisebox{-0.5ex}{\ding{55}} & \raisebox{-0.5ex}{\checkmark}  & \raisebox{-0.5ex}{\ding{55}} & \raisebox{-0.5ex}{\ding{55}} &  \raisebox{-0.5ex}{\checkmark} \\
      \bottomrule
   \end{tabular}
   }
   \label{tab:model_comparison_clip_sam}
\end{table}

\section{Method Discussion}
\label{sec:more_relatd_work}

\subsection{Comparison with Recent Joint SAM and CLIP Models} Several recent works~\cite{chen2023semantic,zhang2023recognize,wang2023sam,han2023boosting} also explore joint segmentation and recognition as one system.
\textbf{Recognize Anything}~\cite{zhang2023recognize} adopts a paradigm for image tagging. The focus of that work is to build large-scale tagging datasets via automatic text semantic parsing. The SAM model is only an external module. Therefore, it cannot perform well with box or mask-level recognition~\cite{zhang2023recognize}. 
\textbf{SAM-CLIP}~\cite{wang2023sam} integrates multi-task learning, continual learning techniques, and
teacher-student distillation, where its goal is to build a single model capable of segmentation and classification. However, its classification capability is limited to the image level since it does not provide optimization for small objects and falls short of the original CLIP. Conversely, our Open-Vocabulary SAM fuses knowledge from SAM and CLIP via SAM2CLIP and CLIP2SAM, which keeps the original CLIP and can support interactive segmentation, instance-level classification, and scale-up beyond the frozen CLIP.
\textbf{SSA}~\cite{chen2023semantic} also builds a system that integrates the knowledge from CLIP and SAM. It involves a semantic voting branch for refining the semantic segmentation results generated by an external semantic segmentation model and refined by SAM. On the contrary, Open-Vocabulary SAM does not require the external segmentor and takes visual prompts rather than coarse semantic segmentation results as input. Moreover, Open-Vocabulary SAM focuses on instance segmentation rather than semantic segmentation.
\textbf{Sambor}~\cite{han2023boosting} builds on frozen SAM and CLIP, but adds a SideFormer to fuse the knowledge. The SideFormer serves as an adapter for generating object proposals. Compared to Open-Vocabulary SAM, however, the design keeps two foundation models, which introduces extra computational costs similar to combined baselines. In addition, Sambor does not perform interactive segmentation like Open-Vocabulary SAM.

\begin{figure*}[t]
    \centering
   \includegraphics[width=1.\textwidth]{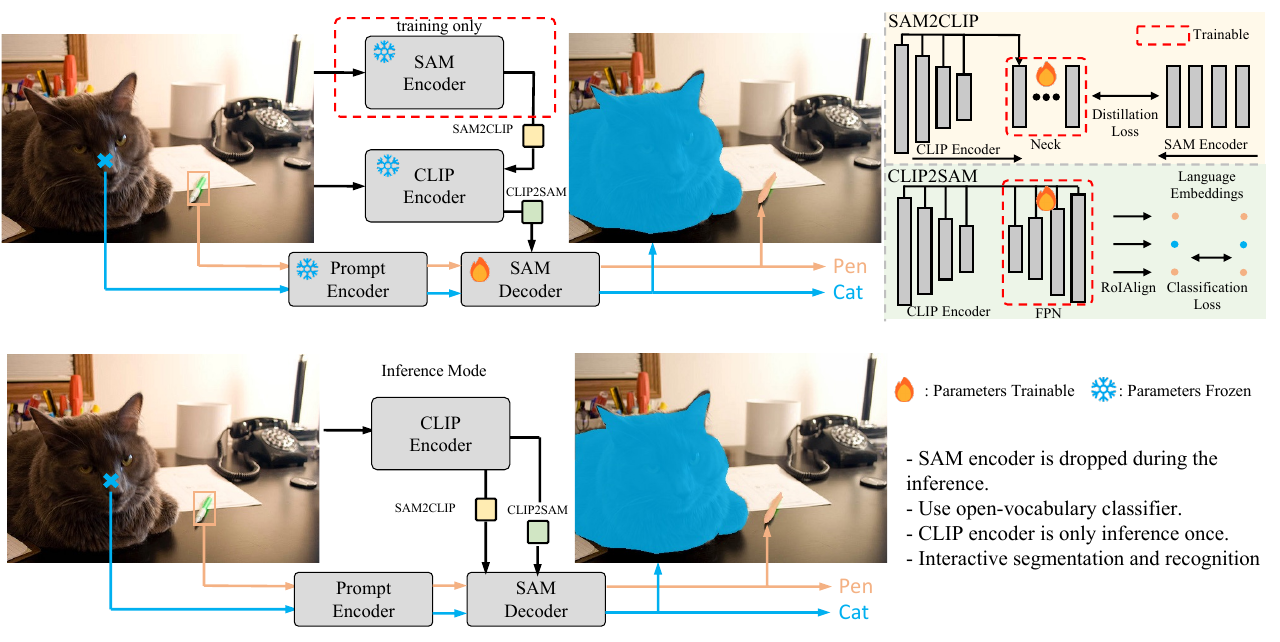}
    \caption{Illustration of Open-Vocabulary SAM. For training, the SAM encoder is adopted as a teacher network, while SAM2CLIP plays the role of a student network and aligns the knowledge of SAM into CLIP. The CLIP2SAM transfers the CLIP knowledge to the SAM decoder and performs joint segmentation and classification for close-set and open vocabulary settings. For inference, we drop the SAM encoder and only take the CLIP encoder. Each visual prompt is only encoded and decoded by the prompt encoder and lightweight SAM decoder, which is the same as SAM.}
    \label{fig:method_supp}
\end{figure*}

\subsection{Comparison with Open-Vocabulary Methods} 
Although our method can recognize objects in an open-vocabulary setting, our work is in a different setting compared to previous open-vocabulary methods.
Specifically, we aim to build a SAM-like model with interactive prompts, such as points and boxes. 
On the contrary, previous open-vocabulary detection methods~\cite{zhou2022detecting,gu2021open,zang2022open,zareian2021open} need to generate region proposals or mask proposals to recall all possible objects in the scene. 
Thus, most approaches aim to improve the recall of novel objects in region proposal networks or mask proposal networks~\cite{zareian2021open,cheng2021mask2former}.
Meanwhile, previous open-vocabulary semantic segmentation methods~\cite{xu2022simple,wu2023betrayed,wu2023clipself,wu2023towards} focus on cross-dataset evaluation mode. 
Our Open-Vocabulary SAM is different from all these works.
Our model, following the same spirit of SAM, mainly takes visual proposals for segmentation and recognition. 
Thus, not only can users deploy our model to take the visual prompts from humans but also deploy our model on various OV-detectors~\cite{zhou2022detecting, wu2023clipself} to achieve instance segmentation or class-agnostic detectors to achieve joint segmentation and recognition.
Moreover, once scaled up with massive datasets, our method is more practical than previous open-vocabulary works mainly designed for several specific datasets, such as ADE-20k, Pascal-VOC, or LVIS. We provide an interactive demo that can take visual prompts from humans to demonstrate real-world performance.

\subsection{Training and Inference Details} 
In Fig.~\ref{fig:method_supp}, we present a more detailed visualization of both training and inference of our Open-Vocabulary SAM. As shown in the figure, only three components are learned during the training, including SAM2CLIP (129M), CLIP2SAM (3.8M), and SAM decoder (4.0M). The remaining parameters are frozen. The total parameters are 304M. After SAM2CLIP distillation, the heavy SAM encoder (637M) is dropped when fine-tuning our model on the detection and classification datasets, which speeds up the CLIP2SAM process. 
The CLIP2SAM module is co-trained using a diverse mixture of datasets, encompassing detection, segmentation, and classification tasks. 
Specifically, for segmentation, the training integrates both mask and classification labels, while for detection and classification tasks, the focus is solely on training the classification head. 
The classification dataset, notable for its extensive range of class categories, coupled with the inclusion of several recent works that feature an extensive vocabulary, significantly enhances the model's capabilities. 
After this extensive co-training process, our model exhibits the remarkable ability to recognize and segment over twenty thousand class categories, all within a single, unified framework.

\begin{table*}[t!]
    \centering
    \caption{Comparison of mask quality with various detectors on LVIS dataset. We report the mask mean AP for comparison. The masks are generated by each method, while the labels are from the corresponding detectors.}
    \resizebox{\linewidth}{!}{
    \begin{tabular}{cc|ccc|ccc|ccc|cc}
    \toprule[0.2em]
    Method & Detectors & mAP & AP50 & AP75 & APS & APM & APL & APr & APn & APf & \#Params & FLOPs \\
    \midrule
    SAM-Huge &  Detic (swin-base) & 43.5&59.8&46.5&31.4&55.3&60.6&42.6&43.3&44.2 & 641M & 3,001G  \\
    Open-Vocabulary SAM  & Detic (swin-base) & 42.6&59.5&45.1&29.7&53.6&61.6&42.0&42.5&42.9 & 304M & 1,180G \\
    \hline
    SAM-Huge &  ViTDet (Huge) & 44.5&61.5&47.4&34.2&56.1&60.4&34.6&45.4&47.8 & 641M & 3,001G  \\
    TAP-L~\cite{pan2023tokenize} & ViTDet (Huge) & 42.6 & -- & -- &29.8&55.5&64.8&33.3&43.6&45.5 & 312M & 1502G\\
    Open-Vocabulary SAM  & ViTDet (Huge) & 44&61.7&47.1&33&54.9&61.6&34.5&44.9&47.3 & 304M & 1,180G \\
    \bottomrule[0.2em]
    \end{tabular}
    }
    \label{tab:mask_qual_lvis}
\end{table*}

\begin{figure}[t]
    \centering
   \includegraphics[width=\textwidth]{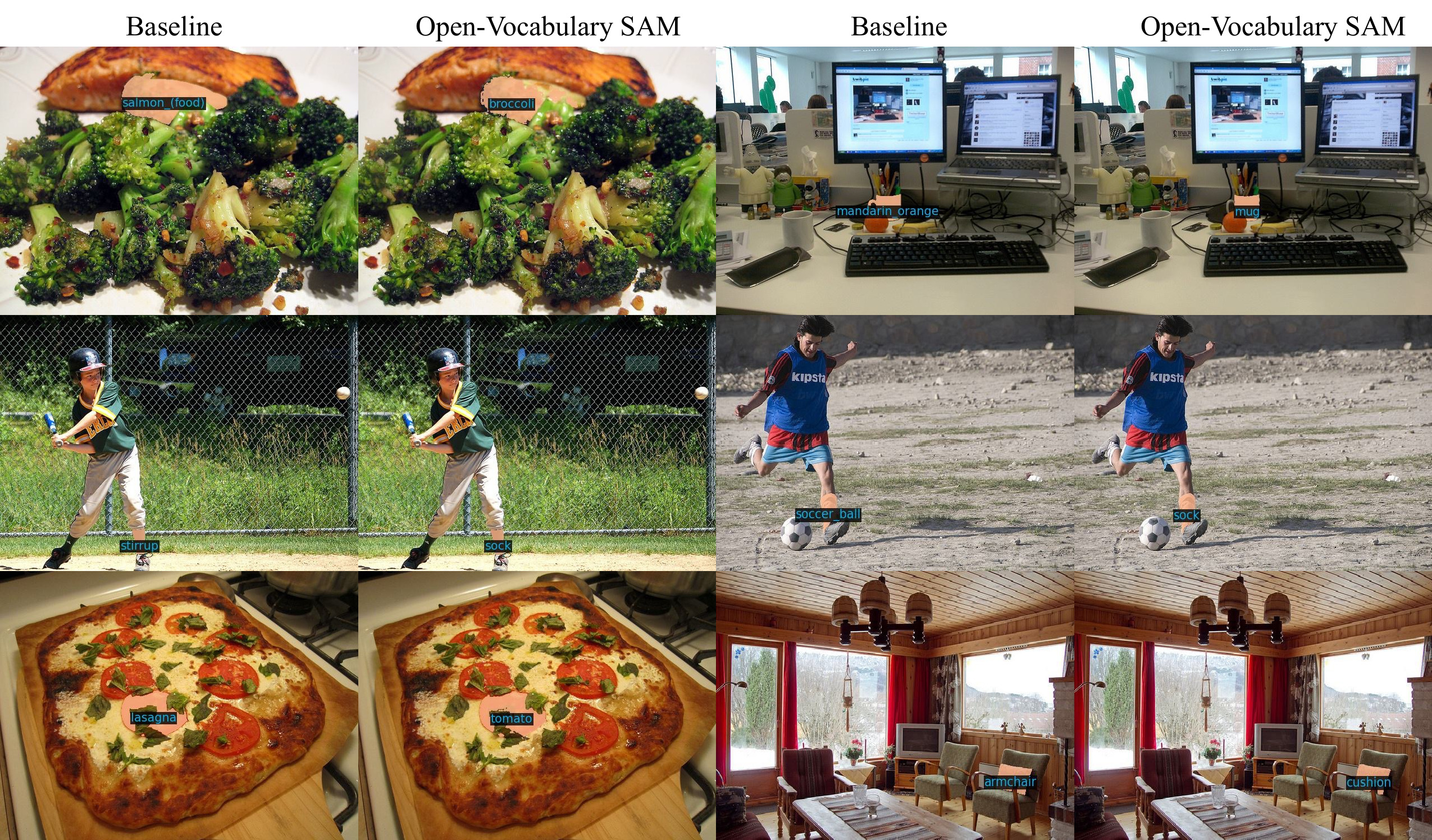}
    \caption{Visualization comparison of Open-Vocabulary SAM and the feature-crop baseline. We print the class name output by the model on the corresponding masks.}
    \label{fig:vis_comp}
\end{figure}
\begin{figure}[t]
    \centering
   \includegraphics[width=\textwidth]{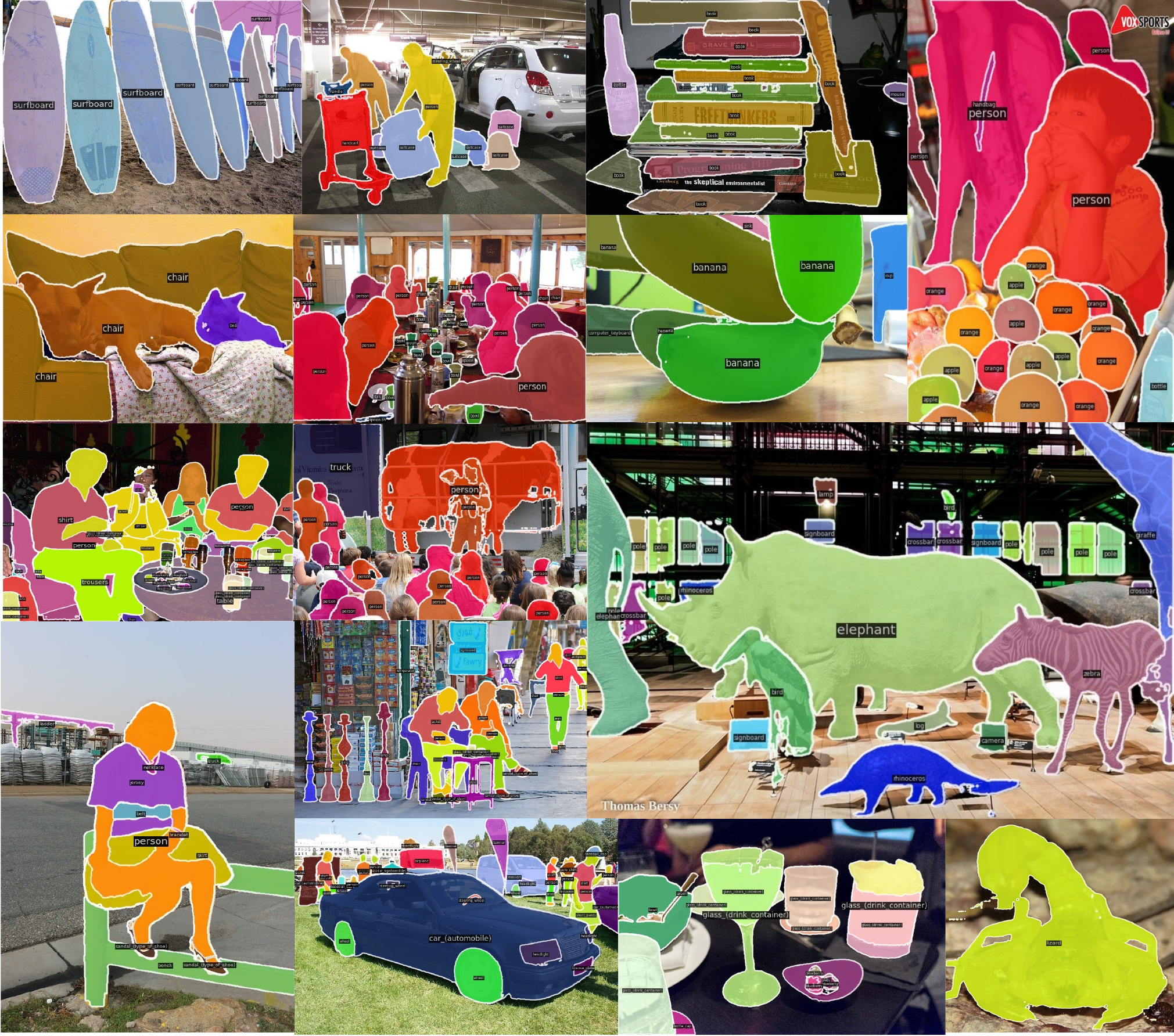}
    \caption{Visualization of everything mode.}
    \label{fig:everything}
\end{figure}
\begin{figure}[t]
    \centering
   \includegraphics[width=\textwidth]{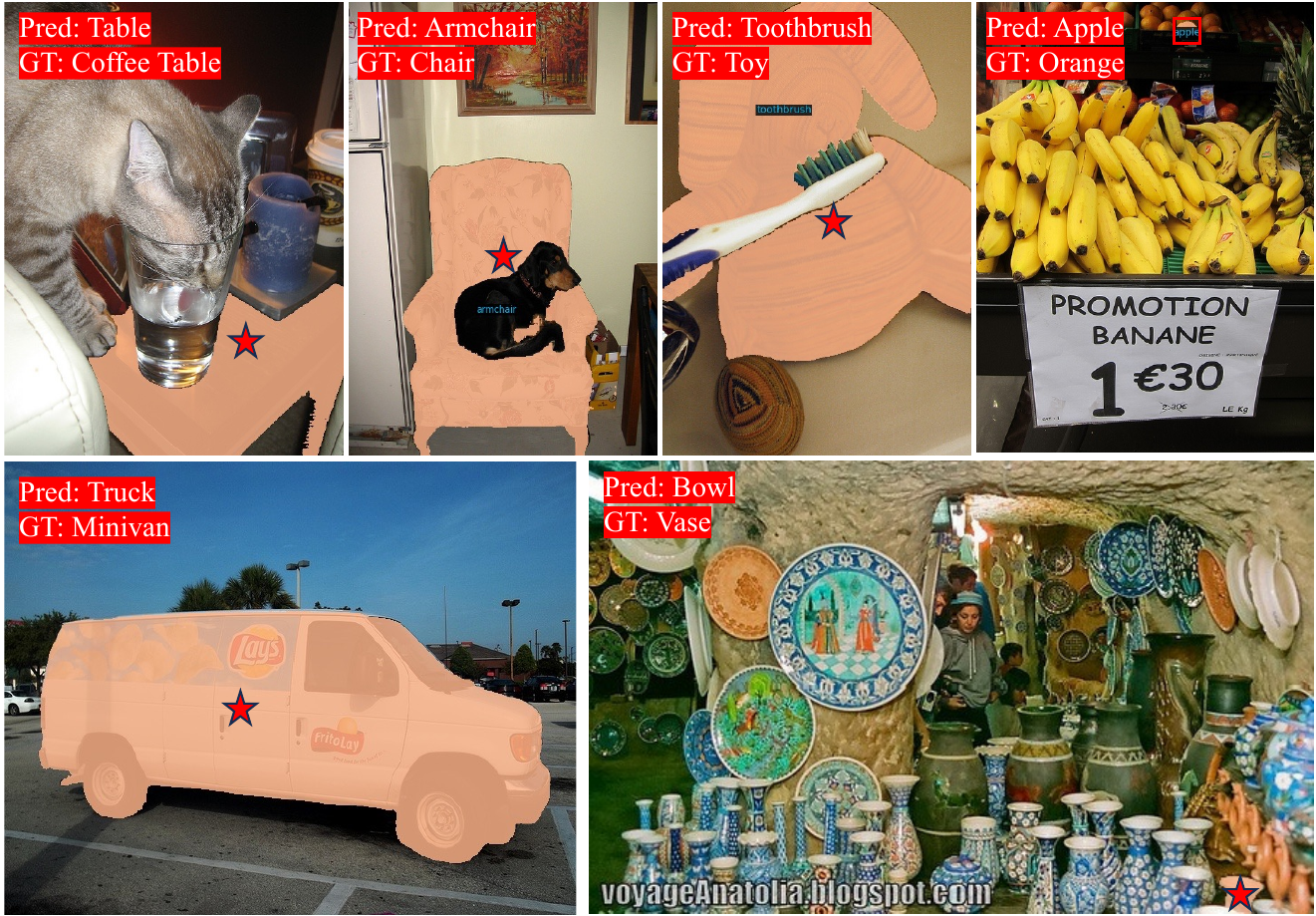}
    \caption{Failure cases of Open-Vocabulary SAM.}
    \label{fig:failure}
\end{figure}
\begin{figure*}[t]
    \centering
   \includegraphics[width=\textwidth]{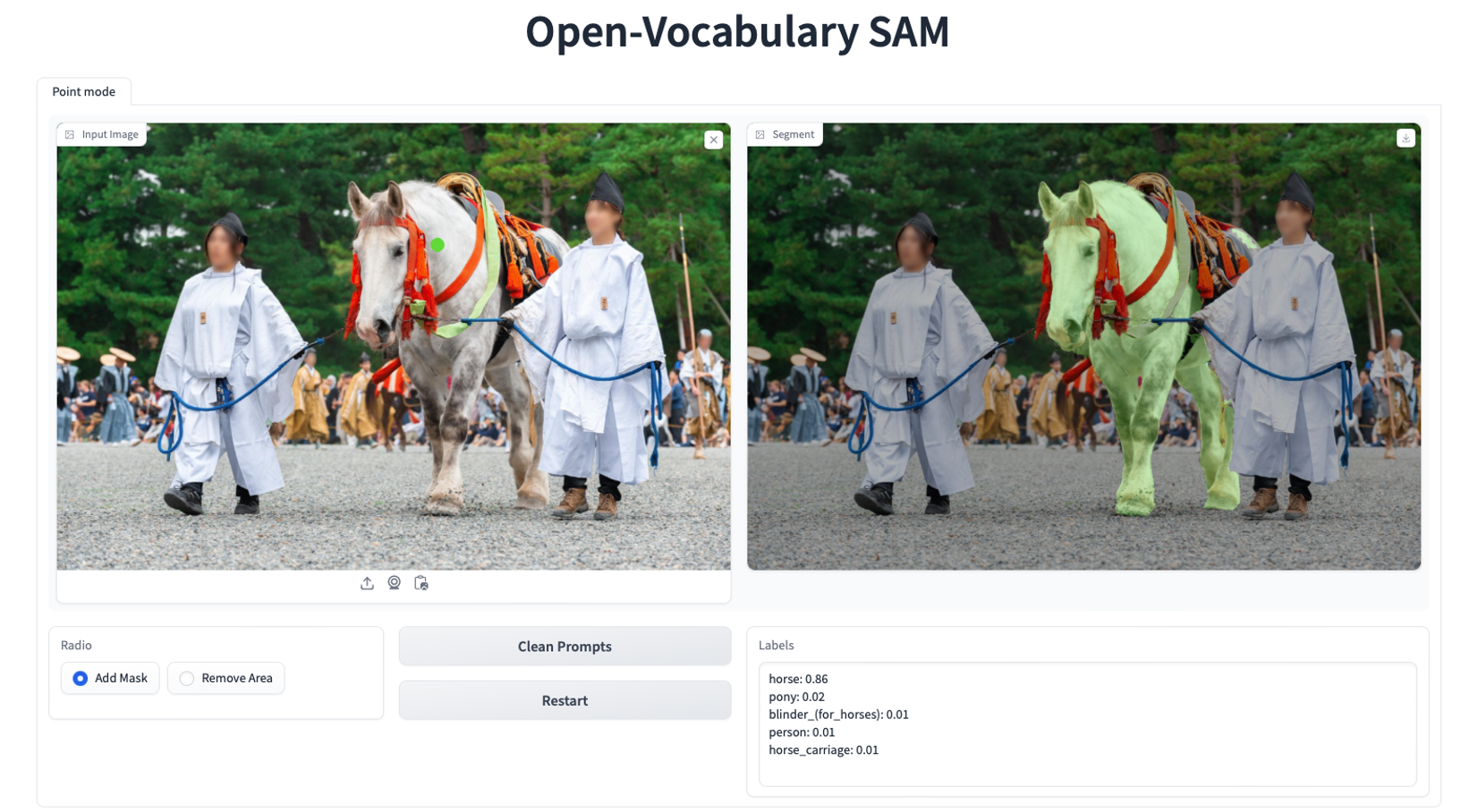}
    \caption{With a simple click, our Open-Vocabulary SAM can generate the mask and label of the object. Please refer to \textbf{tools\_demo.mov} for the interactive demo.}
    \label{fig:ovsam_tool}
\end{figure*}

\section{More Experimental Results}
\label{sec:more_exp_results}

\subsection{More Comparison with SAM} In the main paper, we already compare our method with SAM on the COCO dataset (Tab.4), which shows that Open-Vocabulary SAM has good performance compared to the original SAM. In Tab.~\ref{tab:mask_qual_lvis}, we further compare our method with SAM and a recent SAM-like method TAP~\cite{pan2023tokenize}. The results of TAP-L~\cite{pan2023tokenize} are from the original paper. Regarding the mask quality, we notice that Open-Vocabulary SAM has comparable or slightly worse results compared to the original SAM. However, we want to note that Open-Vocabulary SAM has a much lower computational cost. When comparing with recently proposed TAP~\cite{pan2023tokenize}, Open-Vocabulary SAM has better performance and lower computational cost.

\noindent
\subsection{More Visualization Comparison} In Fig.~\ref{fig:vis_comp}, we compare our method with the feature-crop baseline with visualization results. We noticed that although the recognition capability of the feature-crop baseline is significantly poor compared to Open-Vocabulary SAM, especially on the small objects, the errors are relatively traceable. We noticed that the feature-crop baseline can extract useful information from features, but it may be confused with adjacent objects. We speculate that this may be because it only performs image-level training and lacks object-level recognition capabilities. In contrast, Open-Vocabulary SAM has great advantages at object-level recognition, especially in the recognition of small objects.

\noindent
\subsection{Segment and Recognize Everything}
In Fig.~\ref{fig:everything}, we demonstrate demonstrate the capabilities of Open-Vocabulary SAM on \textit{segment and recognize everything}. With the box prompts as input, our method can segment every identifiable object within a given image, showcasing its versatility and effectiveness in handling diverse objects. Based on the results presented in Fig.~\ref{fig:everything}, Open-Vocabulary SAM not only excels in segmenting a wide range of objects but also in recognizing them, which shows its potential as a powerful tool for comprehensive image analysis.

\section{Failure Cases, Demo and Future Work}
\label{sec:demo}

\subsection{Failure Cases} In Figure~\ref{fig:failure}, we present a series of instances where our Open-Vocabulary SAM does not perform well. First, Open-Vocabulary SAM struggles to differentiate between classes that are subtly distinct from one another, such as mistaking "Truck" for "Minivan". %
Moreover, some challenging cases, such as extreme occlusion, also confuse Open-Vocabulary SAM. For example, Open-Vocabulary SAM incorrectly classified ``Vase'' as ``Bowl'' in the bottom right example. Besides these challenging scenarios, another notable and interesting issue is the model's confusion with overlapping objects. It misclassified the ``Toy'' to the ``Toothbrush'' (the third example in the figure), where the two objects have significant overlapping. 
These examples highlight areas where further research could be beneficial in enhancing the performance of Open-Vocabulary SAM.

\subsection{Short Introduction To Demo} We present an online demo in addition to our paper. It shows the functionality of our Open-Vocabulary SAM (please also refer to Fig.~\ref{fig:ovsam_tool} for illustration), which can segment and recognize various classes on many scenes. Please refer to our GitHub repo at \url{https://github.com/HarborYuan/ovsam} for more details.

\subsection{Future Work} While users can efficiently interact with specific objects using point-and-click or box-dragging techniques, future work will explore using coarse masks or language descriptions as interactive prompts. We aim to continue investigating these promising new directions. Also, our work mainly focuses on foreground categories (e.g., person), and future work may consider introducing background categories (e.g., sky) for comprehensive segmentation and recognition.

\noindent
\textbf{Acknowledgements.} This study is supported under the RIE2020 Industry Alignment Fund-Industry Collaboration Projects (IAF-ICP) Funding Initiative, as well as cash and in-kind contributions from the industry partner(s). The project is also supported by Singapore MOE AcRF Tier 1 (RG16/21) and the National Key R\&D Program of China (No. 2022ZD0161600).

%
%
\bibliographystyle{splncs04}
\bibliography{ref}
\end{document}